# The PICCO Framework for Large Language Model Prompting: A Taxonomy and Reference Architecture for Prompt Structure

David A. Cook, MD, MHPE

*Mayo Clinic College of Medicine and Science | Mayo Clinic Division of General Internal Medicine*
*Rochester, MN, USA*

ORCID: https://orcid.org/0000-0003-2383-4633 | cook.david33@mayo.edu

## Abstract

Large language model (LLM) performance depends heavily on prompt design, yet prompt construction is often described and applied inconsistently. Our purpose was to derive a reference framework for structuring LLM prompts. This paper presents PICCO, a framework derived through a rigorous synthesis of 11 previously published prompting frameworks identified through a multi-database search. The analysis yields two main contributions. First, it proposes a taxonomy that distinguishes prompt frameworks, prompt elements, prompt generation, prompting techniques, and prompt engineering as related but non-equivalent concepts. Second, it derives a five-element reference architecture for prompt generation: Persona, Instructions, Context, Constraints, and Output (PICCO). For each element, we define its function, scope, and relationship to other elements, with the goal of improving conceptual clarity and supporting more systematic prompt design. Finally, to support application of the framework, we outline key concepts relevant to implementation, including prompting techniques (e.g., zero-shot, few-shot, chain-of-thought, ensembling, decomposition, and self-critique, with selected variants), human and automated approaches to iterative prompt engineering, responsible prompting considerations such as security, privacy, bias, and trust, and priorities for future research. This work is a conceptual and methodological contribution: it formalizes a common structure for prompt specification and comparison, but does not claim empirical validation of PICCO as an optimization method.

**Keywords**: Prompt Engineering; Prompting Techniques; PICCO Framework; LLM Optimization; Artificial Intelligence; Human-Computer Interaction; Natural Language Processing; Medical Education; Instructional Design; Educational Technology

## Introduction

Large language models (LLMs) put sophisticated artificial intelligence (AI) within reach of nearly everyone.[1, 2] However, optimal use of these tools requires a new skillset – namely, how to write an effective prompt ("the input to an [LLM] that is used to guide its output"[3]). Although several books (e.g., these[4-6]) and numerous online resources (e.g., these[7-9]) address prompt engineering, there remains a need for a concise, evidence-informed framework to support practical prompt design. Systematic reviews have cataloged applications and techniques of prompt engineering[3, 10-14] but these largely target system developers rather than end users. One tutorial focused on writing prompts for clinical queries;[15] however, broader guidance across tasks and contexts is needed. A concise, rigorously derived, user-centered prompting framework would enable educators, clinicians, and other non-experts to use these tools effectively and responsibly.

Moreover, prompt frameworks (general blueprints for structuring a prompt) are widely advocated to ensure complete, effective prompts; yet, as shown below, existing frameworks lack scholarly grounding. A well-grounded prompt framework would promote prompts that yield outputs aligned with users' intents.

The purpose of this study is to derive and describe a rigorously developed framework for prompt generation, clarify related concepts in prompt design, and outline key considerations relevant to implementation. The principles apply directly to text generation and analysis; multimodal tasks (audio, image, video) may require adaptation. Most examples draw from health professions education, but the principles apply broadly across fields including education, clinical medicine, business, computer programming, and personal life.

## Background

### Key terms

Table 1 contains definitions for terms related to prompt engineering. Historically, many of these terms have been used loosely and interchangeably; greater consistency and precision in vocabulary will facilitate communication and teaching. In particular, the term "prompt engineering" is commonly misused. Strictly speaking, prompt engineering refers to deliberate, iterative *refinement* of a prompt.[3] "Prompt *generation*" is a better label for the process of *composing* a prompt, which is what most users do when interacting with a chat-based LLM such as ChatGPT. Additionally, users often seek to optimize reusable prompt *templates* that can be scaled across multiple data inputs or repeated tasks.

# Table 1. Key terms related to prompt engineering

| Term | Definition |
|------|-----------|
| ***Prompting and prompt frameworks*** | |
| Prompt | "Input to a Generative AI model that is used to guide its output. Prompts may consist of text, image, sound, or other media instructions submitted to the LLM."[3] |
| Prompting | Process of providing a prompt to an LLM. |
| Prompt template | Pre-defined pattern of words with placeholders that can be filled to create a complete prompt. In this example, "<u>Please score the following essay from 1 to 10</u>: [text]," the underlined words are the prompt template; the template + actual essay text together would constitute the full prompt. |
| Prompt generation (development, writing, creation) | Process of composing a prompt. |
| Prompt framework | General blueprint for structuring a prompt or prompt template to facilitate completeness and effectiveness (i.e., the component parts or elements of a complete prompt). |
| Prompt element | Constituent part of a prompt framework. |
| Prompt engineering | Process of deliberately, iteratively evaluating and refining a prompt to optimize performance. |
| Prompting technique | Systematic method for structuring or sequencing a prompt or sequence of prompts to enhance performance (e.g., including an example of the desired output). |
| Prompt engineering technique | Systematic method for conducting prompt engineering (i.e., iteratively refining a prompt); can be manual or automated. |
| In-context learning | a) The general ability of LLMs to infer (learn) how to respond using only information contained within a prompt, without task-specific training or programming.<br>b) [narrower usage] The inclusion of 1 or more examples (input-output pairs) within a prompt; synonymous with the prompting technique of few-shot prompting (see Table 3). |
| ***Prompt features, tools, and strategies*** | |
| Delimiter | Characters or words that mark a segment of the prompt.<br>This segment might comprise text to be analyzed (input data or context), specific steps to follow, or the beginning of a new section. For example, instructions could request the LLM to "Translate the text starting after ###." Delimiters help both the LLM and humans understand the prompt and obtain desired output by increasing precision, reducing ambiguity, and providing structure. Delimiters can be 1 or more unique characters (\|, ''', $$$, ZZZZ), command words ([TRANSLATE]), brackets (<...>, {...}), or tags (a descriptive word in angle brackets <> marking the start and [prefaced by "/"] the end of a section, e.g., <text>...</text>). It usually helps to define the delimiter in the instructions, although sometimes the LLM can infer the intent (e.g., starting a new section using ###, or enclosing an essay using tags <essay> ... </essay>). |
| Hyperparameter | A setting specified outside the training process that shapes how the LLM generates or scores text, such as randomness, creativity, output length, or decoding strategy. Hyperparameters relevant to typical end-users include temperature, top-P, max tokens, max output length, and penalties that discourage repetition (e.g., frequency or presence penalty). Some contemporary LLMs allow model-specific hyperparameters such as reasoning level. |

| Term | Definition |
|---|---|
| Temperature | A hyperparameter that controls the randomness of output by reshaping the token probability distribution.<br><br>A low temperature produces a tall, narrow distribution, concentrating probability on a small set of high-likelihood tokens; as a result, the LLM selects the same tokens more often, making responses more deterministic, predictable, and reproducible. A higher temperature flattens and widens the distribution, increasing the chance of sampling lower-probability tokens and making responses more random. This is good if the goal is creative responses (e.g., poetry), novel ideas, or diverse answers across multiple replications (e.g., qualitative data analysis looking for innovative themes); however, very high temperatures often result in incoherent text and factual inaccuracies. A typical default temperature is 0.7. |
| Top-P (nucleus sampling) | A hyperparameter that controls the diversity of a response by adjusting the proportion of tokens eligible for selection.<br><br>Conceptually it is like slicing off the top of the probability distribution. Slicing off only the very tip (e.g., top-P 0.10) provides a small set of highly probable tokens, resulting in focused and predictable output. Slicing much deeper (e.g., top-P 0.95) provides a large set of tokens with wide variability, resulting in diverse and creative (but potentially less relevant) output. A typical default top-P is 0.9. Top-k is a related hyperparameter that regulates the *number* instead of the proportion or eligible tokens; it is rarely used today. |
| Fine-tuning | The process of further training a pretrained LLM using a curated set of input-output pairs (labeled or annotated data) so it can perform a specific task with high proficiency.<br><br>Since the LLM starts with extensive pre-trained knowledge, fine-tuning can be done using far fewer exemplars (as few as 15 for simple tasks) compared with training a model from scratch. Although there is an up-front cost to collect annotated data and complete the fine-tuning, the fine-tuned model can give high quality, high efficiency, highly-tailored responses. Fine-tuning involves repeated training cycles or ***epochs***, each composed of several training steps that analyze a subset (*batch*) of the training dataset. Key fine-tuning hyperparameters include the number of epochs, ***batch size*** (number of exemplars per step), and ***learning rate*** (how much the model adjusts from one step to the next). |
| Retrieval Augmented Generation (RAG) | A method in which an LLM retrieves relevant information from a separate knowledge base and incorporates that information into its response.<br><br>LLMs can fail to give accurate or complete responses due to random errors (hallucinations) or domain-specific gaps in their training data. RAG addresses both limitations by grounding responses in an external knowledge base. The LLM interprets the user request, retrieves information from a trusted (e.g., bespoke or vetted public) knowledge base, and then generates a fluent, accurate, up-to-date, context-aware response. The quality of the final answer depends heavily on the quality of the knowledge base and retrieval strategy. |

| Term | Definition |
|---|---|
| Plug-in | External software tool that extends the LLM's capabilities by enabling specialized functions. |
| | LLMs are limited in their ability to perform non-text operations such as exact math, statistical analyses, and interacting with external systems. Plug-ins enable these and other tasks (e.g., scheduling a meeting, planning travel, searching a proprietary database, composing an email, or summarizing documents). Some LLMs allow direct access to the plug-in, enabling the LLM to invoke the tool automatically. More often, the LLM is prompted to generate a structured query that the user submits to a specific plug-in; the plug-in's output can then be used directly, or fed back into the LLM for further processing. Plug-in availability varies by platform, and not all LLMs have the same capabilities. For example, plug-ins must be explicitly enabled in ChatGPT or configured through the OpenAI GPT API. |
| Agent | Advanced software system that used LLMs in conjunction with other computer tools to autonomously plan, sequence, and complete multi-step tasks. |
| | An agent can decompose a task into component steps, plan and sequence these steps, identify and use tools to accomplish each step, and monitor its progress toward the overall goal. Examples include agents supporting automated data analysis workflows, literature reviews, virtual patient simulations, customer service (e.g., chatbots), email triage, and travel planning. |
| | Plug-ins are tools that extend LLM capabilities; agents are autonomous, goal-directed systems that use these and other tools to plan and execute multi-step processes. |
| Context document (Persona document) | Document that can be attached to every prompt addressing related tasks, containing information about the audience, setting, goals, historical background, specialized knowledge, and desired persona or tone (see Table 3 for details and examples of what might be included). |

Abbreviations: AI, artificial intelligence; LLM, large language model.

## How LLMs work

LLMs are often described as probability machines: they are trained using vast amounts of data to recognize patterns in text, images, and sound, and then to generate new text, images, and sound based on probabilities. Imagine the possible (probable) completion of a sentence such as "To be or not to be, that is the ___." Most readers, recognizing this as part of Hamlet's soliloquy, would choose the word "question." Phrases such as "conundrum," "dilemma," or "existential tension" would all be plausible – but not very probable. Irrelevant words such as "humanity," "quickly," or "purple"; or even random letters ("alkjnan") could also be used – but these would be implausible and thus highly improbable. LLMs generate responses by repeatedly computing a probability distribution for the next term ("token") given all prior tokens, then selecting a token from that distribution – step-by-step, 1 token at a time.

Why does this matter? First, text that is probable or plausible can still be wrong. This causes the common problem of "hallucinations." Here, the LLM generates a pattern of words that probably belong together, but result in an incorrect statement (e.g., a factual error, historical inaccuracy, logical inconsistency, or math mistake). For example, Geoff Norman – an education scientist with hundreds of publications on clinical reasoning – might *plausibly* have published an article entitled "Etiology of Errors in Clinical Reasoning: System 1 vs System 2" in the journal *Academic Medicine*, but he did not.

Second, the purpose of a prompt is to increase the likelihood of certain terms in the response and decrease the likelihood of others. This explains why telling the LLM *not* to do something will sometimes paradoxically increase the likelihood that it does this, because the presence of a key word or phrase increases the likelihood of a response that incorporates the unwanted term.

Third, the generated output includes a degree of randomness (it is "stochastic"). This contrasts with most computer operations, which give an identical solution with each repetition ("deterministic").

Fourth, the response quality depends on the training data. LLMs merely mimic the data they were trained on. If training data are biased or inaccurate, the most probable response will reflect those biases or inaccuracies.

Finally, LLMs rely on surface-level semantics even when they appear to be reasoning logically.[16] Seemingly minor changes in the prompt (e.g., phrasing, sequence, punctuation) can materially influence output.[17-23] Emerging approaches – including several discussed below – attempt to mitigate this.[24-27]

Most current LLMs are also limited by the date of their training data. An LLM trained on 1 November 2024 would not know the current US president. Tools are emerging to circumvent this limitation.

## Why prompts matter

Prompt quality matters: LLM responses are sensitive to subtle choices in wording or structure. Some very simple prompts work adequately. For example, the query, "Who was the 14th president of the United States?" will usually generate the correct response (Franklin Pierce). Other simple prompts comprise a completed example (input-output pair) followed by a new input without output, such as "Night: Day. Fast: ___" (response: "Slow"); or "Name: Nombre. Beach: ___" (response: "Playa"). The LLM inferred (learned) the intended task ("provide antonym," "provide Spanish translation") using only the example, i.e., in-context learning.

However, most prompts must be far more detailed. Asking a human, "Tell me how to make a peanut butter sandwich," illustrates the importance of detailed instructions. People often omit "obvious" steps like "get a knife" or "open the jar," yielding instructions that, if followed literally, lead to failure. Similarly, the LLM prompt template "Analyze this reflective essay and provide a score and feedback: [essay]" is likely to generate inconsistent responses; more details and more structure are needed. A prompt framework – a general blueprint for structuring a prompt – can help to ensure completeness and increase the probability of receiving the desired response.[28-30]

# A rigorously derived prompt framework

Several prompt frameworks have been proposed (see Table 2), but most have obscure origin and none detail a scholarly derivation of their constituent elements. For example, the earliest prompt framework is CRISPE, which (per Internet lore) was introduced internally by OpenAI staff and then diffused externally through peer networking and online dissemination. Despite CRISPE's wide use, sources document its structure but not its derivation.[31] For many popular frameworks the original source is a YouTube video,[32, 33] blog,[34] or enterprise website.[7, 35]

To remedy the lack of a rigorously derived framework, I systematically identified and synthesized all available sources into a novel framework.

## Methods

### Identification of existing frameworks

I identified existing prompt frameworks by a) searching the Internet using a search engine and 2 LLMs (a comprehensive but non-systematic approach), and b) systematically searching peer-reviewed publications (reproducible, but much narrower body of literature).

I started with an unstructured query using Google. Building from this, I separately queried OpenAI GPT-4.1 and Gemini-2.5-flash using several iteratively evolving prompts (multiple sessions, 2 October to 6 November 2025). These searches identified 8 unique prompt frameworks[7, 9, 31-36] (Table 2). I also documented the earliest description of each framework. I included frameworks with obscure origin (e.g., CRISPE) if they were cited in ≥2 sources (e.g., publications, websites), suggesting acceptance in the prompt engineering community. The LLMs identified several additional "state-of-the-art frameworks" that ultimately proved to be hallucinations (novel amalgamations of best practices with no verifiable prior use).

Building from these findings and with support from an experienced research librarian, on 5 November 2025 I conducted a systematic search of Scopus, Web of Science, Embase, and Preprint Citation Index using key terms including large-language-model, LLM, prompt, framework, and 4 specific frameworks (CRISPE, COSTAR, RICCE, RICECO). I made no restrictions on language. This yielded 143 records. I selected for inclusion all publications (N=3) that described a specific prompt framework (Table 2).[28-30] Most excluded studies described fully-developed task-specific prompts (N=76), novel prompting techniques (N=31), and novel prompt engineering approaches (N=25). The full search strategy and inclusion/exclusion details are reported in Appendix Table A1.

### Synthesis

I first enumerated the constituent elements and operational definitions of each framework, and identified repeatedly-shared elements (i.e., cross-framework convergence). I next considered the sequence of elements in each framework, again searching for cross-framework convergence. The final sequence integrated additional published evidence.[17, 19, 22, 25]

# Table 2. Derivation of the PICCO framework: Cross-framework synthesis of existing prompt frameworks

| Framework (year) PICCO (novel) | Framework elements[a] | | | | | |
| --- | --- | --- | --- | --- | --- | --- |
| | Persona | Instructions | Context | Constraints | Output | --[b] |
| CRISPE (~2022)[31c,d] | Capacity (1) Role (2) Personality (5) | Statement (4) | Insight (3) | | | Experiment (6) |
| COSTAR (2023)[7c] | Style (3) | | Context (1) Objective (2) Audience (5) | Tone (4) | Response (6) | |
| RICCE (2023)[35c,d] | Role (1) | Instruction (2) | Context (3) | Constraints (4) | | Examples (5)[e] Evaluation (5)[e] |
| LangGPT (2024)[30f,g] | Profile Skills | Initialization Workflow Suggestions | Goals Background | Constraints | Style Output format | Examples |
| PERFECT (2024)[36c] | Role (3) | Element (2) Examples (5)[h] Timeframe (6) | Purpose (1) | Conditions (5) | Format (4) | |
| RACE (2024)[34c] | Role (1) | Action (2) | Context (3) | | Execute (4) | |
| RISEN (2024)[32c] | Role (1) | Instruction (2) Steps (3) | | Narrowing (5) | End goal (4) | |
| 5C (2025)[28f] | Character (1) | | Cause (2) | Constraint (3) Contingency (4) | | Calibration (5) |
| LearnPrompting (2025)[9c] | Role (3) | Directive (4) | Additional information (2) | | Output formatting (5) | Examples (1) |
| Promptomatix (2025)[29f,g] | | Task Instructions Rules | Context Question | | Output format | Examples Tools |
| RICECO (2025)[33c,d] | Role (1) | Instruction (2) | Context (3) | Constraints (5) | Output (6) | Examples (4) |

[a] Elements in the PICCO row were empirically derived from a synthesis across the other frameworks. Elements in other rows are labels from the original (existing) framework. Numbers in parentheses indicate the position of each element in the original framework.

[b] The elements in this column are not part of a prompt framework as defined herein (see Table 1): "Examples" are a prompting technique. "Experiment," "evaluation," and "calibration" are aspects of prompt engineering.

[c] Identified from Internet search using Google, OpenAI GPT-4.1, and Gemini-2.5-flash.

[d] These frameworks have no clear scholarly derivation; they appear to have emerged among AI enthusiasts and are now accepted in the community (cited in ≥2 other publications or websites). For each framework I have cited the earliest source I could find (e.g., YouTube video, blog post, etc.).

 Different sources list slightly different expansions of this acronym: most say "Examples," but some say "Evaluation."
[f] Identified from systematic search of peer-reviewed literature.
[g] There was no explicit position of the elements in this framework.
[h] "Examples" in the PERFECT framework are requested *from* the LLM (i.e., instructions request "examples or case studies, if relevant"), not provided *to* the LLM as part of the prompt.

## Results

### Synthesis of frameworks

Table 2 presents the framework elements and the cross-framework synthesis. Individually, each framework was incomplete (missing at least one element), overinclusive (including "elements" better classified as prompting techniques or prompt engineering techniques), and/or redundant (distinguishing separate elements that could be collapsed into 1). Nonetheless, collectively these readily support a novel 5-element framework: Persona, Instructions, Context, Constraints, Output (PICCO; Table 3).

Several frameworks included Examples as an additional key element. However, I agree with Schulhoff[3] in classifying Examples as a prompting technique (few-shot prompting) rather than a key framework element. The problems with including few-shot prompting – or any other specific prompting technique – as a key element are that in a given situation a particular technique might be unnecessary or even counterproductive; and other techniques might be as or more important.

### Final framework

*Persona*: The first element specifies a role for the LLM,[37] such as "You are a ... physician-educator," "patient with diabetes," "NIH-funded researcher," or "expert Python programmer." Why does this seemingly trivial element make a difference? LLMs have been trained using essentially all data available on the Internet, yielding a huge library of information. Assigning a persona is like selecting a specific book. Using very few words, the persona narrows the LLM's probability distribution to the desired information domain. The persona can optionally specify details such as personality, tone, and expertise. Assigning multiple people (e.g., a mock debate) or a non-human persona ("You are a horse") can generate creative insights.

*Instructions*: The next element specifies what the LLM should do – the central task. Instructions should be explicit, detailed, and succinct. It commonly requires multiple iterations (prompt engineering) to optimize the details, phrasing, and sequence of instructions.[23, 26, 27, 29] Details might include: A specific question or desired end product, specific steps, required elements (e.g., length, rating scale), and specific techniques (see Table 4).

*Context*: The context specifies background information that will make the LLM response relevant, specific, and purpose-aligned. Making analogy with acting, one author explained, "The Instruction is the line you want them to say. The Context is their character's entire backstory, their motivation and the emotional state of the scene."[8] Details might include the purpose or intended application; target audience; setting or situation (user/institution setting, historical information, background statistics or data); specialized knowledge (theory, practice guidelines, business strategies, technical specifications); or knowledge database (e.g., for retrieval-augmented generation[38]). However, research shows that irrelevant context impairs performance.[39]

*Constraints*: Constraints provide guardrails – guidelines, boundaries, and rules on what *not* to do, or crucial requirements that merit repeating. These can include words to avoid, restrictions on length or information sources, security safeguards, and requests to confirm accuracy. Note that instructions on what to *do* are generally preferred over constraints on what *not* to do;[40] constraints should emphasize – not introduce – boundaries.

*Output*: The final element specifies the desired response structure. Table 3 lists numerous possibilities.

As regards the *sequence* of prompt elements, research indicates that this influences LLM output[17, 22, 41, 42] but existing evidence does not support definitive universal guidelines. The PICCO sequence reflects a "majority vote" synthesis across frameworks (Table 2). However, it is unclear whether logic, personal (empirical) experience, or desire for a nice mnemonic determined the sequence in these frameworks. A minority of frameworks put Context ahead of Instructions, suggesting that Persona-*Context-Instructions*-Constraints-Output might be reasonable in some situations. Supporting this idea, one website suggested that "placing the [instructions] last helps avoid the AI continuing the additional information [context] instead of focusing on the task at hand."[9] More consistency across frameworks was found in putting Persona first, and Constraints and Output last. The optimal sequence remains a question open to further research.

## Table 3. Elements of the PICCO framework

| Element | Explanation | Possible details | Examples |
|---|---|---|---|
| Persona | Assign LLM a specific role or personality. | • Identity/role (e.g., doctor, patient, teacher, researcher, librarian, communication expert, poet, Nobel-winning economist, Python programmer)<br>• Experience, expertise, knowledge, skills<br>• Age, country of origin, other demographic<br>• Personality, preferences<br>• Tone (e.g., formal, friendly, sassy, sarcastic, empathetic, kind, humorous, authoritative)<br>• Multiple persons at once (diverse opinions, mock debate) | • You are a board-certified family medicine physician who is really good at translating evidence-based recommendations into informal language that patients can understand.<br>• You are a research librarian with 20 years' experience doing systematic reviews.<br>• You are an innovator who thinks outside the box. Your expertise is to draw insights from diverse fields and apply them to a novel problem or task. |

| Element | Explanation | Possible details | Examples |
|---------|-------------|------------------|----------|
| Instructions | Provide a detailed description of the desired task, action, or product (what LLM should do, phrased in positive terms). | • Task, desired product, question<br>• Essential content, length, structure<br>• Specific actions, steps, or style (but note: LLMs may struggle with multi-step prompts; it often works better to submit different prompts for each step in a multi-step process, rather than bundling them into a single prompt)<br>• Priorities (e.g., comprehensiveness, reasoning, accuracy, speed)<br>• Specific techniques (see Table 4) | • Give me a 150-word explanation of why it is important to control type 2 diabetes.<br>• What is 17 * 7?<br>• I will upload a PDF of a research study. I want you to summarize this in a 250-word paragraph that includes the research question framed using the PICO format, a short description of study methods, 2-4 key findings, and 2 actionable take-home implications. Let's think about this plan step by step before you actually start working. |
| Context | Provide background information (why, where, or for whom this is being done; specialized knowledge or guidelines). | • Purpose (why this is needed, how it will be used)<br>• Target audience (including their needs and expectations)<br>• Background<br>   ○ User or institution setting (location, preferences, mission, culture)<br>   ○ Historical information (the situation or story thus far)<br>   ○ Raw data or statistics (background data to justify need or clarify current state of evidence)<br>• Specialized knowledge<br>   ○ Proprietary information (unpublished data, business intelligence)<br>   ○ Relevant theory or conceptual frameworks<br>   ○ Practice guidelines, industry best practices or standards<br>   ○ Business strategy<br>   ○ Technical specifications<br>• External sources with more information (e.g., Retrieval-Augmented Generation [RAG])<br>• Tone (as per Persona)<br>• If exemplars are used (few-shot | • The target audience is adult patients with diabetes in your outpatient clinic. You want to motivate them to take good care of their diabetes. Use the clinical practice guidelines that I upload as the primary source for information. You may also search the Internet for additional evidence-based practices. Use a relaxed, empathetic tone that targets a 10th-grade reading level.<br>• The target audience is first year family medicine residents. The attached data from your medical center show that patients who see residents have worse control of diabetes than patients who see staff physicians. Use the attached practice guidelines to guide your recommendations. Use a kind but authoritative tone. |

| Element | Explanation | Possible details | Examples |
|---|---|---|---|
| | | learning), they should be included with Context. | |
| Constraints | Prescribe guidelines, boundaries, and rules on what LLM should *not* do, and imperative requirements. | • Crucial requirements for length, structure, steps (re-emphasize instructions)<br>• Words, content, or style to avoid (buzzwords, bias, jargon, speculation)<br>• Promotion of accuracy (avoid fabrication, support claims)<br>• Restriction to specific sources (PubMed-listed citations, peer-reviewed literature, practice guidelines, Facebook posts)<br>• Parameters for safety and security; rules to pre-empt hacking and malicious use (stay in role, stay strictly on topic, refuse inappropriate requests)<br>• Discrete steps to confirm accuracy, verify anticipated trouble spots | • Avoid medical jargon or cliché phrases. Do not use exclamation marks. Be sure the paragraph is less than 200 words. Double-check each recommendation against the practice guidelines.<br>• Make sure each reference comes from a reputable peer-reviewed or respected preprint source and that the citation is real (not a hallucination, no fabrications). Give me the PMID or DOI for each reference. |
| Output | Specify the precise structure of the final product. | • Product structure or format, such as:<br>  ○ Numeric: integer; number between 1 and 100; number formatting<br>  ○ Text: 1 sentence, 150-word paragraph<br>  ○ Structured text: bulleted/numbered list, table, letter, email<br>  ○ Essay: persuasive writing essay, creative essay<br>  ○ Task-specific: teaching plan, workshop outline, research abstract, business brief, data insight, social media tweet<br>  ○ Consumer-oriented: question & answer (FAQ), step-by-step instructions<br>  ○ Structured data: JSON, XML<br>  ○ Computer code: script for Python, JavaScript, R, SQL<br>  ○ Downloadable file: Word doc, CSV file, PDF | • The final product should be a 300-word paragraph.<br>• Frame the output as "frequently asked questions" and a brief response; keep the total length less than 300 words, and put it in a Word doc that I can share with patients.<br>• Frame the output as a 300-word abstract for presentation at a national meeting, using the IMRD format. |

Abbreviations: LLM, large language model.

## Table 4. Prompting techniques: Systematic methods for structuring or sequencing a prompt

| Technique | Description | Design decisions or sub-techniques[a] |
|---|---|---|
| Zero-Shot Prompting | Use a basic prompt comprised of PICCO elements without exemplars or other specific techniques. | • <u>Meta-prompting</u>: ask LLM to rewrite a prompt to add or remove details or refine wording, or to generate an initial prompt.<br>• <u>Role/style prompting</u>: specify the persona, style, tone, genre.<br>• <u>Emotion/persuasion prompting</u>: add text to encourage "best effort" from LLM (e.g., highlight the importance of an accurate response; add social cues such as authority or reciprocity; add a bribe or threat).<br>• <u>Rereading</u>: tell LLM to "Read the question [or instructions] again" before responding. |
| Few-Shot Prompting | Include 1 or more exemplars as part of the prompt (in-context learning[b]). The various sub-techniques represent different ways to create, select, refine, format, and position the exemplars. | • <u>Exemplar selection</u>: select optimal exemplars; considerations include: similar vs divergent from test sample (similar usually better); balance/distribution of classifications (balanced usually better than uneven distribution); human vs AI selection (many AI approaches can automate and optimize selection, e.g., these[24-27]); and quantity of exemplars (more is usually better, but too many exemplars can adversely affect performance[51]).<br>• <u>Exemplar labeling</u>: ensure accurate classification (labeling, rating, annotation) of exemplars (human-generated, LLM-generated [automated] labels) (but note that research suggests that incorrect exemplar labels do not necessarily reduce downstream accuracy[45, 46]).<br>• <u>Exemplar format</u>: present exemplars using optimal format (varies by task).<br>• <u>Exemplar position and order</u>: present exemplars in ideal position (e.g., putting the block of exemplars at the start of the prompt is usually better than in the middle or end[22]) and order (sequence of exemplars within the block[17, 26]) (varies by task).<br>• <u>Instruction selection</u>: optimize instructions to accompany exemplars; when using exemplars, instructions can often be simple and brief (some studies[48, 49] suggest that simple or generic [task-agnostic] instructions are better than more detailed instructions when exemplars are used).<br>• <u>Self-Generated In-context Learning</u>: ask LLM to generate fabricated (simulated) exemplars in a separate, preparatory query. |

| Technique | Description | Design decisions or sub-techniques[a] |
|---|---|---|
| Chain of Thought (CoT) (Thought Generation) | Ask LLM to articulate its reasoning path as part of the response, or to follow the reasoning path explicitly represented in 1 or more worked examples (input-reasoning-output triplets). | • Generic (zero-shot) CoT: add a generic request "Let's think step by step" or "First tell me your approach to solving this problem" or "First, let's think about this logically" before acting on the remainder of the prompt. Variations include longer or more specific instructions to think logically.<br>• Few-shot CoT: include exemplars that model a correct reasoning path (i.e., input-reasoning-output "worked examples" of *reasoning* that LLM can then follow). Variations include showing both correct and incorrect reasoning paths, and different ways to generate the reasoning paths. |
| Ensembling | Generate several responses and then synthesize a final response (select the best single response, or combine responses into a final response). The various sub-techniques represent different ways to generate prompt(s), generate varying responses (by using different prompts or exemplars), or synthesize (select/combine) responses. | • Prompt generation and selection: generate different versions of the prompt, then generate responses for each, then synthesize final response (several approaches described).<br>• Consistency: obtain multiple responses for the same prompt, then synthesize final response (several approaches described).<br>• Synthesis: generate a final response; can systematically identify the single best response or combine several responses into a single response (several approaches described). |
| Decomposition | Break a complex task or question into a series of smaller, simpler sub-tasks; tasks are then completed in sequence (final response is result of last sub-task) or in parallel (final response combines results of each sub-task). The various sub-techniques represent different ways to decompose the task into sub-tasks, label or represent each sub-task, complete each sub-task, and synthesize a final response from sub-task results. | • Task decomposition: use LLM to break the task into sub-tasks (several approaches described).<br>• Sub-task labeling: assign labels or placeholders to each sub-task to facilitate the combining of results at the end (several approaches described).<br>• Sub-task completion: optimize each sub-task; approaches include separately-optimized "handlers" (e.g., LLM prompt, Python script) for specific sub-tasks, using exemplars, and trial-and-error with different prompts for each sub-task.<br>• Sub-task sequencing: define the sequence of sub-tasks (e.g., completed in sequence [one after another] or in parallel [all at the same time]).<br>• Synthesis of sub-task results: generate or select a final response; approaches include using the final sub-task result, evaluating each sub-task result individually and selecting the best, combining sub-task results using explicit logical operators, and using ensemble approaches (above) to combine responses. |

| Technique | Description | Design decisions or sub-techniques[a] |
|-----------|-------------|---------------------------------------|
| Self-Criticism | Ask LLM to confirm or improve the prompt or response by critiquing their own output. The various sub-techniques represent different ways to check the accuracy or quality of the final response or reasoning path. | • <u>Check accuracy of response</u>: use LLM in a separate step to confirm the correctness of the original response (several approaches described).<br>• <u>Improve response</u>: ask LLM for suggestions to improve the response (several approaches described).<br>• <u>Improve prompt</u>: ask LLM for suggestions to improve the prompt or CoT reasoning path (several approaches described). |

[a] See Schulhoff[3] and [learnprompting.org](learnprompting.org) [58] for details and citations for each sub-technique. See Appendix Table A2 for additional details.

[b] Some definitions of "in-context learning" include all situations (including Zero-Shot Learning) in which an LLM "learns" to complete a task from any prompt (as contrasted with requiring task-specific training).

### Including data and exemplars

Many prompts request analysis of text or numeric data, either as a one-shot prompt or as a prompt template with a placeholder for added data. The data to be analyzed should usually be appended at the end of the prompt, or uploaded as a separate file, to clearly distinguish it from Instructions and Context. Exceptions arise when analyzing very long texts; for these, a very brief reminder prompt (PICCO prompt – data – 1-sentence recap) or interleaved/sequential analysis (task-1 prompt – data – task-2 prompt – data) may be appropriate. Best practice guidelines[40] recommend use of labels and delimiters such as "### DATA ###" or "[DATA]" to clearly demarcate data (see Table 1).

Exemplars, if used for few-shot learning, should be included in the Context. Although several prompt frameworks position exemplars at or near the end (see Table 2), research shows that this impairs LLM performance.[19, 22] Appending exemplars at the end may increase the risk of misinterpretation or confusion with target data, particularly in long-context tasks.[9] Instead, few-shot exemplars should be part of the Context, clearly delimited and positioned before output specifications and before the target data.

## Additional prompting strategies and tools

Table 1 describes prompt features and tools that control or improve the LLM response, including delimiters, hyperparameters, fine-tuning, retrieval-augmented generation,[38] plug-ins, and agents.

The PICCO framework – indeed, any prompt framework – primarily helps the human author write a complete and optimally sequenced prompt. It is usually unnecessary to label each element in the prompt.

Uncertainties will inevitably arise in prompt wording, in how information is classified (e.g., Persona, Instructions, or Context) and sequenced, and in the selection of prompting techniques. In situations where optimal prompt performance is essential, such ambiguities should be resolved empirically through rigorous, iterative prompt engineering as outlined below.

Users often find themselves repeatedly writing similar prompts (e.g., to perform a literature search or statistical analysis, or to generate an email, lesson plan, or marketing strategy). To maintain consistency and reduce redundant effort, a well-written prompt element or prompt template (i.e., a combination of several elements) can be saved, and then pasted or attached to future prompts.[8] Some LLMs offer "Custom GPTs" or other features that enable similar functionality.

LLMs often struggle with complicated multi-step prompts. To mitigate this, clearly delimit and label each goal and/or interim step. ("This task will comprise 4 distinct steps. Complete each step entirely before proceeding to the next. ### Step 1. ..... ### Step 2. ..... [etc.]"). An alternative (and most robust) solution would break each step into separate prompts.

## Prompting techniques

Prompting techniques are blueprints for creating or structuring a prompt or sequence of prompts to enhance its effectiveness. An insightful systematic review[3] identified 6 classifications of prompting techniques (Table 4, and more details in Appendix Table A2).

### Zero-shot prompting

The most basic technique – *zero-shot prompting* – asks a (pretrained) LLM to generate a response without task-specific examples. Beyond using a robust prompt framework (as above), several sub-techniques can further improve zero-shot performance.

One of the most powerful techniques is *meta-prompting*, which asks the LLM to help write or improve the prompt. Contemporary LLMs excel in helping users craft effective prompts.

Some studies have found that motivation cues can improve LLM performance; these include appeals to an authority figure, praise ("You're so impressive"), reciprocity ("I spent time preparing this information for you"), and unity ("We want to ...").[43] Other work suggests that emphasizing the importance of accuracy ("It is crucial that you get this right"), or adding threats or bribes ("I'll pay you a million dollars if you get this right"), can enhance performance, although more recent research contradicts these findings.[44] Another technique simply instructs the LLM to "Re-read the question [or instructions]."

### Few-shot prompting

*Few-shot prompting* adds 1 or more input-output exemplars to the prompt. The input typifies the text or non-text content to be analyzed. The output closely models the ideal response, both content and format. Several frameworks include "Examples" as an element, but – as proposed by Schulhoff[3] – this is best viewed as an optional (albeit effective) technique.

This technique is often called *in-context learning* because the LLM "learns" from examples included in the prompt ("in-context") rather than a separate training phase. However, in-context learning also refers broadly to an LLM's ability to "learn" to respond to any prompt without task-specific training or programming, which includes zero-shot prompting. Since both usages are correct, I avoid using in-context learning to refer solely to few-shot prompting.

Design decisions in few-shot prompting focus on how to create, select, refine, format, and position the exemplars, including the number and variety of exemplars, methods to ensure accurate labeling, and how to present input-output pairs. All of these remain active topics of ongoing investigation.[17, 19, 21, 22, 25-27, 42, 45-51] As explained above, exemplars should be included with the Context (never at the end). Research further shows that when multiple exemplars are used, their order influences responses, although there are no generalizable rules.[17, 25, 26, 42]

Incorrect exemplar labels do not necessarily reduce downstream accuracy;[45] rather, evidence suggests that exemplars probably serve primarily to illustrate input/output structure, format, and task framing.[45, 46] Thus, when real human-labeled exemplars are few, an LLM can generate exemplars for subsequent use (Self-Generated In-context Learning). One study found that fabricated exemplars were inferior to real exemplars but better than none.[47]

## Chain of thought prompting

The term *chain of thought (CoT) prompting* refers to 2 distinct techniques that both focus on the LLM's reasoning path (thought generation). *Generic (zero-shot) CoT* simply asks the LLM to "think step by step" or "explain your approach before giving the answer."[46, 52] Many contemporary "reasoning" LLMs now do this internally without asking, but it can still be helpful to ask the LLM to articulate its reasoning to troubleshoot and optimize the prompt.

*Few-shot CoT* uses exemplars that explicitly model a correct reasoning path (i.e., input-*reasoning*-output triplets).[53] Although early studies demonstrated substantial improvements with few-shot CoT, benefits with contemporary LLMs may be smaller because these models already perform strong internal reasoning.[16, 46, 49, 54]

## Ensembling

*Ensembling* encompasses a variety of prompting approaches that a) generate multiple responses and b) select or synthesize a final response. Some variations focus on the generation stage (e.g., multiple replications using the same prompt or a paraphrased prompt). Other variations focus on the selection stage (e.g., rigorous approaches to select the single best response or combine several responses into one).

## Decomposition

*Decomposition* encompasses various approaches to systematically break a complex task into a series of smaller sub-tasks. Variations focus on how tasks are decomposed, how sub-tasks are completed, and how sub-task results are combined into a final response.

### Self-criticism

*Self-criticism* is an extension of meta-prompting that asks the LLM to critique its own output and use this critique to improve the prompt or revise the output. Variations focus on how the LLM critiques the output (e.g., appraising quality, identifying errors), and how it uses this information (to refine the prompt or to further refine an already-generated response).

## Prompt engineering

Prompt engineering is the process of iteratively improving a prompt (e.g., cycles of Build [a draft prompt]-Measure [evaluate]-Learn [revise];[55] or Plan-Do-Study-Act[56]). Proposed approaches include human appraisal, meta-prompting (asking the LLM to critique the prompt or output), and automated optimization systems, each with several specific techniques.[3]

Human appraisals evaluate the prompt (identifying assumptions, omissions, clarifications, and irrelevant or unnecessary words) and the output (accuracy, completeness, format, and alignment with goals).[7, 8, 57, 58] Meta-prompting asks the LLM to do the same.[3] Several automated optimization systems have been described that evaluate outputs using an objective quality criterion and iteratively refine a prompt until it achieves a specified level of performance;[3, 24, 25] Appendix Table A3 lists 25 publications identified in the literature search that describe such systems.

Following both human and automated appraisals, prompts can be iteratively refined through systematic paraphrasing, targeted additions or deletions, and using a second prompt to critically appraise the first.

## Discussion

This article presents a framework and taxonomy for prompt generation derived from prompting literature and practice. While the proposed Persona-Instructions-Context-Constraints-Output (PICCO) framework has not been formally tested, it reflects rigorous synthesis across 11 expert-proposed frameworks that show increasing convergence over time. Notably, the newest framework – RICECO[33] – differs from PICCO only in treating exemplars (few-shot prompting) as a framework element rather than a technique. Assuming this convergence reflects accumulating practical wisdom, the PICCO framework reasonably captures a current consensus view of prompt structure.

Prompt engineering is evolving rapidly, as is our understanding of how LLMs generate responses and how frequent, user-opaque model updates reshape LLM behaviors, including instruction-following. In my experience, newer LLMs (e.g., GPT-5.2) sometimes prioritize what they infer the user *wants* over what the user explicitly *instructs*, resulting in markedly suboptimal results, while older models (GPT-4o) demonstrate better compliance. As LLMs evolve, prompt engineering will remain a moving target, influenced by task- and model-dependent trade-offs among instruction fidelity, autonomous reasoning, predictable responses, and alignment with human intent.

Although reasoning algorithms and meta-prompting can offload some aspects of prompt generation, the human-in-the-loop remains essential. Skilled prompt engineers integrate understanding of LLM mechanics, domain-specific knowledge and vocabulary, clear communication, and creativity to translate complex goals into precise prompts.

## Limitations

Derivation of the PICCO framework involved only 1 investigator. Given the decentralized nature of this field, a truly comprehensive search is likely impossible; however, extensive searching across web sources, LLMs, and multiple literature databases makes it unlikely that any salient frameworks were missed. Only 3 of the 11 prompt frameworks in Table 2 originate from peer-reviewed literature, and most lack robust derivation; nonetheless, they appear to be adopted in the prompt engineering community. Although the source frameworks were individually limited, their systematic synthesis represented the strongest feasible approach to creating a rigorously grounded framework.

## Responsible prompting

Responsible prompting requires attention to *security*, which concerns how LLMs might mishandle information or be manipulated; and *alignment*, which concerns whether the LLM response matches the user's intent.[57, 59-63] Even well-crafted prompts can fail when LLMs behave unpredictably or reproduce flaws in their training data. The most responsible approach is human-AI collaboration[64] in which users apply consistent fact-checking and critical judgment to decide when to trust, refine, or override AI outputs. Table 5 elaborates on these issues.

## Research on prompting

Prompting is an active, dynamic field of research, and the questions proposed above represent only a small fraction of current evidence gaps. High quality, relevant investigations are desperately needed.

Importantly, the generalizability of research on prompting is limited in at least 2 ways. First, LLMs are evolving rapidly. Research on last year's LLMs may not apply to those now available, and today's findings may not apply to the next generation. Findings may also fail to generalize across LLM families with different architectures or tuning (OpenAI GPT, Gemini, Claude, Llama, or GPT-4.1 vs GPT-o4).

Second, research is highly task- and context-specific. The optimal prompt or prompting technique might differ substantially depending on the goal – such as providing factual answers, performing math, executing complex reasoning, analyzing a free-text short answer or 1-page essay, generating Python code, or translating language. It may also depend on contextual factors such as domain (academia, medicine, business, law, personal use), stakes (high, low, or casual), and whether external tools or plug-ins are enabled.

Moreover, prompting research is commonly published in preprints rather than slower peer-reviewed journals; and much knowledge exists as experiential wisdom available only through blogs and social media, if at all. While perhaps inevitable in a fast-moving field, this puts the burden of research appraisal on the end-user rather than peer review. It also underscores the need to conduct and disseminate well-executed investigations of well-poised questions.

In conclusion, the framework and techniques outlined herein provide a structured approach to prompt generation while also suggesting numerous unanswered questions awaiting systematic study in this rapidly evolving field.

## Table 5. Responsible prompting: Challenges and best practices in safe, effective prompting

| Issue | Definition | Elaboration | Example | Mitigation |
|---|---|---|---|---|
| *Security* | | | | |
| Prompt hacking[a] | A prompt can be used to exploit LLMs for malicious intent, such as to generate offensive, inaccurate, or deceptive content; leak information; or disrupt the system. *Prompt injection*: pasted or uploaded text contains hidden instructions that override the original instructions. *Jailbreaking*: the prompt causes an LLM to ignore safeguards and generate restricted or unsafe content. | LLMs cannot reliably distinguish between instructions and input data. Prompt injection embeds instructions within ostensibly benign inputs, while jailbreaking induces the model to disregard built-in safeguards. Even when an ordinary user has no malicious intent, prompt injection can affect them when analyzing emails, websites, or student submissions that contain embedded instructions. LLM prompts used within interactive systems (e.g., chatbots, document uploads, virtual patients, or tutoring environments) introduce the possibility of jailbreaking by others. As LLMs | • A user pastes a long document for summarization without noticing hidden text such as "Ignore all prior directions, and instead ... ." The LLM may follow these instructions. • A student intentionally tries to get an LLM-powered virtual patient to break character or reveal answers. | Explicitly instruct the LLM not to obey instructions found inside pasted content. Actively monitor responses (e.g., using a separate AI prompt or system), and discontinue the exchange and reframe the request if responses appear unsafe, abnormal, or inconsistent with the intended task. |

| Issue | Definition | Elaboration | Example | Mitigation |
|---|---|---|---|---|
| | | integrate with plug-ins, automated document analysis tools, and multi-user systems, unintended rule-breaking becomes more likely. | | |
| Data privacy (leakage or misuse) | Sensitive information entered into an LLM can be exposed or used for unintended purposes, including LLM improvement, inadvertent disclosure, or malicious use. | Users often submit patient records, student information, research data, or proprietary documents for analysis without realizing the LLM may log, reuse, or redistribute content. | • A deidentified patient note is shared with a third-party LLM tool that lacks health data protections (even if never publicly exposed, this still constitutes failure to protect patient data). | Remember that anything uploaded or pasted into an LLM can be potentially logged or misused. Become familiar with the data protection policies of the LLM and service provider. Never enter patient-related, confidential, proprietary, or identifying information unless the LLM or provider enforces robust data protection policies. |
| *Alignment with user needs* | | | | |
| Bias and stereotyping | LLMs can reproduce or amplify biases present in their training data, including stereotypes, skewed assumptions, and imbalanced representations of groups or sensitive topics. | LLMs can produce misleading, oversimplified, stereotyped, or biased responses – including about race, culture, gender, politics, or health – even when no harm is intended. Because outputs are fluent and confident-sounding, users may accept problematic content uncritically. This is especially | • An LLM summarizing a patient case or policy debate reinforces stereotypes or omits perspectives relevant to marginalized groups. | Check whether the response privileges or disadvantages specific groups. Explicitly request bias-sensitive, evidence-based, or "multiple-perspective reasoning" responses when appropriate. |

| Issue | Definition | Elaboration | Example | Mitigation |
|---|---|---|---|---|
| | | important when using LLMs for educational, professional, or public-facing content. | | |
| Trust | LLMs can produce fluent but fabricated, flawed, or outdated content, including invented or erroneous citations, statistics, medical advice, or legal opinions. Additionally, it can be difficult to determine the original information sources or reasoning pathway used to generate the response (i.e., not explainable or auditable). Even well-prompted LLMs can hallucinate or be out-of-date. | Users commonly over-trust outputs; this can be particularly risky in high-stakes domains like medicine, education, and law. Proprietary LLM training limits transparency and accountability, and the reasons for a given response are often not "explainable." | • An LLM provides a reference that looks real but does not exist.<br>• An LLM response contains a direct quote from an unacknowledged source, resulting in inadvertent plagiarism. | Treat LLM output as a draft, not an authority. Verify citations, check facts, and confirm accuracy – especially when stakes are high. Ask the LLM to provide sources for key facts, and to explain its reasoning (and then confirm these). Engage the LLM to confirm each citation or fact using a separate prompt. |
| Prompt drift | When a prompt refined using one LLM is used with a different LLM, or when a given LLM changes over time, the prompt performance may change. | Each LLM is trained and implemented using proprietary methods, and LLM behavior will vary across developers, architectures, and version updates. Even minor changes may alter response style, length, correctness, or speed. | • A prompt developed using ChatGPT is now used with Gemini.<br>• OpenAI GPT-4.1 is replaced by GPT-5.1; responses are now slower and more verbose. | Re-evaluate prompt performance whenever changing to a new LLM or adopting a new version of a given LLM. |

Abbreviations: LLM, large language model.

[a] See IBM.com[63] for an excellent discussion of prompt injection and jailbreaking.

## Disclosures

Acknowledgments: The author thanks Michele McGinnis, MSIS, for her help with the literature search, and Christopher J. Mooney, PhD, MPH, MA, for his constructive comments on an early draft of this manuscript.
Funding/Support: None.
Other disclosures: The author used computer tools, including large language models, to identify and summarize relevant scholarly work, and to help in refining limited portions of the manuscript text. However, all of final text reflects original writing by the author.
Ethical approval: Not required.
Previous presentations: None.

# Appendices

## Appendix Table A1. Details of a systematic literature search for prompt frameworks

**Full search strategies**

*Embase*
Embase 1974 to 2025 November 03 (Wolters Kluwer Ovid interface)
Date searched: November 5, 2025
Records retrieved: 38
Publication type: excluded conference abstracts
Search strategy:
((((Exp *Prompt engineering/) AND (framework*.ti,ab,kf)) OR ((prompt*) AND (framework*)).ti) AND (exp *large language model/ OR (large-language-model* OR LLM OR LLMS).ti,ab,kf.)) OR ((CRISPE OR COSTAR OR RICCE OR RICECO) ADJ4 (prompt* OR framework*)).ti,ab,kf.

*Scopus*
Date searched: November 5, 2025
Records retrieved: 40
Publication type: excluded conference papers
Search strategy:
((TITLE-ABS-KEY (large-language-model* OR LLM OR LLMS)) AND (TITLE ((prompt*) AND (framework)))) OR (TITLE-ABS-KEY ((CRISPE OR COSTAR OR RICCE OR RICECO) W/5 (prompt* OR framework))) AND (LIMIT-TO (DOCTYPE , "ar"))

*Science Citation Index*
Science Citation Index Expanded (SCI-Expanded)—1975-present and Emerging Sources Citation Index (ESCI)—2018-present (Web of Science)
Date searched: November 5, 2025
Records retrieved: 36
Search strategy:
((TS=(large-language-model* OR LLM OR LLMS)) AND (TI=((prompt*) AND (framework)))) OR (TS=((CRISPE OR COSTAR OR RICCE OR RICECO) NEAR/5 (prompt* OR framework)))

*Preprint Citation Index*
Preprint Citation Index (PCI)—1991-present (Web of Science)
Date searched: November 5, 2025
Records retrieved: 78
Search strategy:
((TS=(large-language-model* OR LLM OR LLMS)) AND (TI=((prompt*) AND (framework)))) OR (TS=((CRISPE OR COSTAR OR RICCE OR RICECO) NEAR/5 (prompt* OR framework)))

**Inclusion criteria**
- Included: Any publication describing a specific prompt framework, in any language, without restriction by publication date.
- Prompt framework was defined as: A general blueprint for structuring a prompt or prompt template to

facilitate completeness and effectiveness (i.e., the component parts or elements of a complete prompt).

**Trial flow**

Number of unique records, after deduplication: 143

Number excluded, by reason:
- Described fully-developed task-specific prompts, N=76
- Described novel prompting techniques, N=31
- Described novel prompt engineering approaches, N=25
- Not original research, N=7
- Described test of security vulnerability, N=1

**Number included for review: 3**

## Appendix Table A2. Prompting techniques: Systematic methods for structuring or sequencing a prompt (detailed)

| Technique | Description | Design decisions or sub-techniques[a] | Additional / specific sub-techniques[a] |
|---|---|---|---|
| Zero-Shot Prompting | Use a basic prompt comprised of PICCO elements without exemplars or other specific techniques. | • <u>Meta-prompting</u>: ask LLM to rewrite a prompt to add or remove details or refine wording, or to generate an initial prompt.<br>• <u>Role/style prompting</u>: specify the persona, style, tone, genre.<br>• <u>Emotion/persuasion prompting</u>: add text to encourage "best effort" from LLM (e.g., highlight the importance of an accurate response; add social cues such as authority or reciprocity; add a bribe or threat).<br>• <u>Rereading</u>: tell LLM to "Read the question [or instructions] again" before responding. | • <u>Meta-prompting: System 2 Attention</u>: ask LLM to rewrite the prompt to remove extraneous information before acting.<br>• <u>Meta-prompting: Rephrase and Respond</u>: ask LLM to rewrite the prompt to expand details before acting.<br>• <u>Meta-prompting: Self-Ask</u>: ask LLM if it has follow-up questions before acting.<br>• <u>Role Prompting</u>: assign a specific persona.<br>• <u>Style Prompting</u>: specify the style, tone or genre.<br>• <u>Directional Stimulus Prompting</u>: provide targeted hints or cues to steer model behavior (e.g., salient keywords or concepts)<br>• <u>Emotion Prompting</u>: add text that indicates the importance of accurate response.<br>• <u>Persuasion Prompting</u>: add text that would encourage a human to give their best effort:<br>  o <u>Social Cues</u>: include a phrase that conveys authority, prior commitment, praise, reciprocity, scarcity, urgency, widespread acceptance, or unity.[43]<br>  o <u>Bribes and Threats</u>: include a positive ("I will pay you $1 million") or negative ("I will kick a dog") verbal incentive.[44] |
| Few-Shot Prompting | Include 1 or more exemplars as part of the prompt (in-context learning[b]). The various sub-techniques represent different ways to create, | • <u>Exemplar selection</u>: select optimal exemplars; considerations include: similar vs divergent from test sample (similar usually better); balance/ | • Several specific techniques have been proposed to systematically select, label, format, and/or order exemplars using AI and other algorithms or rules;[3] these are rather technical and |

| | | |
|---|---|---|
| select, refine, format, and position the exemplars. | distribution of classifications (balanced usually better than uneven distribution); human vs AI selection (many AI approaches can automate and optimize selection, e.g., these[24-27]); and quantity of exemplars (more is usually better, but too many exemplars can adversely affect performance[51]).<br>• Exemplar labeling: ensure accurate classification (labeling, rating, annotation) of exemplars (human-generated, LLM-generated [automated] labels) (but note that research suggests that incorrect exemplar labels do not necessarily reduce downstream accuracy[45, 46]).<br>• Exemplar format: present exemplars using optimal format (varies by task).<br>• Exemplar position and order: present exemplars in ideal position (e.g., putting the block of exemplars at the start of the prompt is usually better than in the middle or end[22]) and order (sequence of exemplars within the block[17, 26]) (varies by task).<br>• Instruction selection: optimize instructions to accompany exemplars; when using exemplars, instructions can often be simple and brief (some studies[48, 49] suggest that simple or generic [task-agnostic] instructions are better than more detailed instructions when exemplars are used).<br>• Self-Generated In-context Learning: ask | domain-specific, and beyond the scope of this overview. |

| | | | |
|---|---|---|---|
| | | LLM to generate fabricated (simulated) exemplars in a separate, preparatory query. | |
| Chain of Thought (CoT) (Thought Generation) | Ask LLM to articulate its reasoning path as part of the response, or to follow the reasoning path explicitly represented in 1 or more worked examples (input-reasoning-output triplets). | • <u>Generic (zero-shot) CoT</u>: add a generic request "Let's think step by step" or "First tell me your approach to solving this problem" or "First, let's think about this logically" before acting on the remainder of the prompt. Variations include longer or more specific instructions to think logically.<br>• <u>Few-shot CoT</u>: include exemplars that model a correct reasoning path (i.e., input-reasoning-output "worked examples" of *reasoning* that LLM can then follow). Variations include showing both correct and incorrect reasoning paths, and different ways to generate the reasoning paths. | • <u>Generic (zero-shot) CoT</u> (generic request "Let's think step by step" before acting on the remainder of the prompt): variations include:<br>  o <u>Thread-of-Thought (ThoT)</u>: similar to generic CoT with a longer prompt: "Walk me through this context in manageable parts step by step, summarizing and analyzing as we go." (better for large, complex contexts)<br>  o <u>Step-back Prompting</u>: ask LLM a generic high-level question about relevant facts before acting on the prompt<br>  o <u>Analogical Prompting</u>: ask LLM to generate exemplars that include CoT (similar to Self-Generated In-Context Learning, above)<br>• <u>Few-shot CoT</u> (input-reasoning-output triplets that model a correct reasoning path): variations include:<br>  o <u>Contrastive CoT</u>: include both correct and incorrect reasoning paths<br>  o <u>Reasoning path selection</u>: use a systematic approach to generate or select paths; various approaches have been described using both AI and humans (uncertainty-routed CoT, complexity-based CoT, memory-of-thought)<br>  o <u>Auto-CoT</u>: use zero-shot CoT to generate fabricated (simulated) |

| | | | reasoning paths, and then use these as few-shot CoT exemplars |
|---|---|---|---|
| Ensembling | Generate several responses and then synthesize a final response (select the best single response, or combine responses into a final response. The various sub-techniques represent different ways to generate prompt(s), generate varying responses (by using different prompts or exemplars), or synthesize (select/combine) responses. | • <u>Prompt generation and selection</u>: generate different versions of the prompt, then generate responses for each, then synthesize final response (several approaches described).<br>• <u>Consistency</u>: obtain multiple responses for the same prompt, then synthesize final response (several approaches described).<br>• <u>Synthesis</u>: generate a final response; can systematically identify the single best response or combine several responses into a single response (several approaches described). | • <u>Self-Consistency</u>: submit the identical prompt (with non-zero temperature) multiple times; record each response and select the best response based on majority vote (only works for problems with discrete response e.g., math, logic, multiple choice).<br>• <u>Universal Self-Consistency</u>: submit identical prompt multiple times (same as Self-Consistency); submit all responses to LLM to select the most consistent response based on majority *consensus* (works for open-ended problems).<br>• <u>Max Mutual Information</u>: a systematic approach for selecting a good prompt: create various prompts (however you like) and generate responses for each prompt using test-sample input; calculate a mutual information score for each prompt; select the prompt with highest score.<br>• <u>Mixture of Reasoning Experts</u>: creates several LLM "expert systems" each optimized for different problem (e.g., math, factual response, multi-step reasoning); submits the problem to each expert, then uses an "answer selector" to identify the best response (option to abstain [no response] if confidence is low).<br>• <u>Prompt Paraphrasing</u>: use LLM to create multiple versions of a prompt (e.g., directly request LLM to paraphrase, or translate the prompt into another language then back to English); then submit each prompt; then combine responses or use an algorithm (e.g., |
| | | | Max Mutual Information) to select the best prompt. |
| | | | • <u>Demonstration Ensembling</u>: a variation of few-shot prompting: create several prompts, each with a different set of exemplars; combine responses to synthesize final response (various options). |
| | | | • <u>Consistency-based Self-adaptive Prompting</u>: a robust extension of Auto-CoT (above): submit the identical prompt multiple times (same as Self-Consistency) and record the reasoning path of each response; select the most consistent and most diverse paths to use as exemplars; use this prompt (and exemplars) with Self-Consistency technique to generate final response. |
| | | | • <u>Universal Self-Adaptive Prompting</u>: Similar to Consistency-based Self-adaptive Prompting, but generates and evaluates fabricated (simulated) exemplars instead of reasoning paths, and uses these in few-shot prompting (i.e., an ensembling approach to Self-Generated In-context Learning). |
| | | | • <u>Multi-Chain Reasoning</u>: another extension of Auto-CoT (above), a 3-stage prompt: 1) ask LLM to break the task into several sub-tasks; 2) ask LLM to generate multiple reasoning paths + responses for each sub-task; 3) ask LLM to merge all reasoning paths + responses into a final response. |
| | | | • <u>Diverse Verifier on Reasoning Step</u> (DiVeRSe): focuses on the reasoning path: create various prompts (however you like), submit each prompt multiple times and record the |

| | | | reasoning path, score each path on likelihood of correctness, and then verify each reasoning path step-by-step (requires up-front definition of correct reasoning patterns). |
|---|---|---|---|
| Decomposition | Break a complex task or question into a series of smaller, simpler sub-tasks; tasks are then completed in sequence (final response is result of last sub-task) or in parallel (final response combines results of each sub-task). The various sub-techniques represent different ways to decompose the task into sub-tasks, label or represent each sub-task, complete each sub-task, and synthesize a final response from sub-task results. | • <u>Task decomposition</u>: use LLM to break the task into sub-tasks (several approaches described).<br>• <u>Sub-task labeling</u>: assign labels or placeholders to each sub-task to facilitate the combining of results at the end (several approaches described).<br>• <u>Sub-task completion</u>: optimize each sub-task; approaches include separately-optimized "handlers" (e.g., LLM prompt, Python script) for specific sub-tasks, using exemplars, and trial-and-error with different prompts for each sub-task.<br>• <u>Sub-task sequencing</u>: define the sequence of sub-tasks (e.g., completed in sequence [one after another] or in parallel [all at the same time]).<br>• <u>Synthesis of sub-task results</u>: generate or select a final response; approaches include using the final sub-task result, evaluating each sub-task result individually and selecting the best, combining sub-task results using explicit logical operators, and using ensemble approaches (above) to combine responses. | • <u>Prompt chaining</u>: break a task into smaller sub-tasks, each completed in sequence and building on the output of the previous sub-task; the last sub-task response is the final response.<br>• <u>Decomposed Prompting</u> (DecomP): a subtype of prompt chaining in which each sub-task is completed by a separately-optimized "handler" (LLM prompt, Python script, etc.) that can be re-used across tasks (e.g., with other prompts; "modular" development).<br>• <u>Plan-and-Solve</u>: ask LLM to "understand the problem and devise a plan to solve it" and "then carry out the plan and solve the problem step-by-step;" essentially a zero-shot CoT prompt with 2 explicit steps.<br>• <u>Least-to-Most Prompting</u>: 2-stage prompt: 1) ask LLM to break a task into smaller sub-tasks without completing them, often using exemplars showing how to decompose the task; 2) ask LLM to complete each sub-task in sequence, often using exemplars showing how to complete each sub-task.<br>• <u>Understand-Plan-Act-Reflect (UPAR)</u>: aims to mirror human reasoning (Kantian philosophy) by guiding LLM to: 1) Understand – answer 4 questions about relevant entities, constraints, and relationships; 2) Plan – propose a solution (similar to generic CoT); 3) Act – execute the plan; 4) Reflect – check answers and correct errors. |

- Other techniques are designed to address specific problems:
  - Problem: Some tasks have an unclear solution, and require creative planning or trial-and-error. Solution: <u>Tree-of-Thoughts</u>: 2-stage prompt that decomposes the reasoning process into intermediate reasoning states ("thoughts") and explicitly explores multiple alternatives: 1) "propose" prompts ask LLM to generate potential partial solutions; 2) "value" prompts evaluate each partial response and guide LLM along the optimal path; this approach is computationally inefficient and operationally complex, but valuable for difficult open-ended reasoning tasks.
  - Problem: it takes a long time to generate responses for multi-step problems. Solution: <u>Skeleton-of-Thought</u>: a 2-stage prompt: 1) ask LLM to generate a skeleton of the answer (e.g., 5 key points); 2) ask the elaborate on each element of the skeleton *in parallel* (i.e., concurrently, using a different prompt for each response), thus shortening the overall response time.
  - Problem: LLMs do not always correctly apply logic to link sub-task results (especially important for complex rule-based logical or mathematical tasks). Solution: <u>Chain-of-Logic</u>: combine sub-

| | | | |
|---|---|---|---|
| | | | task results using explicit logical operators ("and", "or"); this technique focuses on the logical links between each reasoning step. |
| | | | o  Problem: LLMs propose a reasoning path, but then fail to follow this path because sub-task results are not correctly retrieved for inclusion in final synthesis (i.e., they are "unfaithful" to their own logic). Solution: <u>Faithful CoT</u>: the CoT exemplar includes labels attached to the result of each sub-task in the reasoning path, and uses explicit logic to combine these labels (sub-tasks) to generate the final response; since each sub-task can be "faithfully" solved and labeled, and the logic linking each label can be "faithfully" followed, the final response is trustworthy and interpretable. |
| | | | o  Problem: LLMs are bad at simple math, and often fail at relatively simple math tasks. Solution: <u>Program-of-Thoughts</u>: asks LLM to generate computer code (e.g., Python script) to solve computational tasks, then uses reasoning to put these together into a final response. |
| | | | o  Problem: Many problems combine both math and language. Solution: <u>Chain-of-Code</u>: combines the computer coding of Program-of-Thoughts with LLM language-based ("semantic") reasoning, to get the best of both |
| | | | techniques. |
| | | | <ul><li>○ Problem: Some tasks exceed LLM context window. Solution: <u>Recursion-of-Thoughts</u>. As LLM works through a complex task it generates placeholders for each complicated sub-task, solves each placeholder task using a separate prompt (i.e., overcoming the context window limitation), inserts this solution into the original prompt, and then continues.</li></ul> |
| Self-Criticism | Ask LLM to confirm or improve the prompt or response by critiquing their own output. The various sub-techniques represent different ways to check the accuracy or quality of the final response or reasoning path. | <ul><li><u>Check accuracy of response</u>: use LLM in a separate step to confirm the correctness of the original response (several approaches described).</li><li><u>Improve response</u>: ask LLM for suggestions to improve the response (several approaches described).</li><li><u>Improve prompt</u>: ask LLM for suggestions to improve the prompt or CoT reasoning path (several approaches described).</li></ul> | <ul><li><u>Self-Calibration</u>: 2-stage prompt: 1) generate response; 2) feed the question + proposed response back to LLM and ask if answer is correct.</li><li><u>Self-Refine</u>: iterative approach: 1) generate response; 2) feed response back to LLM and ask for suggestions to improve; 3) make adjustments, and repeat until no further suggestions.</li><li><u>Chain-of-Verification (CoVe)</u>: 4-stage prompt: 1) generate response; 2) feed the question + proposed response back to LLM and ask it to generate a series of verification questions that test the correctness of response; 3) ask LLM each verification question; 4) use responses from (3) to generate final response.</li><li><u>Self-Verification</u>: 3-stage prompt: 1) generate multiple responses using the same CoT prompt; 2) ask LLM to convert each question + response into a single statement and hide part of original question; 3) ask LLM to predict the hidden part. The response from step 1 that matches with correct prediction in step 3 is the final answer.</li></ul> |

| | | | The difference between Chain-of-Verification and Self-Verification is that the former asks about the correctness of the response, whereas the latter asks about a hidden part of the original question. |
|---|---|---|---|
| | | | • <u>Reverse-CoT</u>: 4-stage prompt, iterative approach: 1) generate initial response including reasoning path; 2) ask LLM to use the reasoning path + response to reconstruct the original question; 3) ask LLM to compare the original vs LLM-reconstructed questions; 4) if important differences are detected, ask LLM to revise the reasoning path and then repeat. |
| | | | • <u>Cumulative Reasoning</u>: 3-stage prompt: 1) ask LLM to propose a reasoning path; 2) ask LLM to verify each reasoning step (and make corrections as needed); 3) after all steps have been verified, generate final response. |

ᵃ See Schulhoff[3] and learnprompting.org [58] for details and citations for each sub-technique.
ᵇ Some definitions of "in-context learning" include all situations (including Zero-Shot Learning) in which an LLM "learns" to complete a task from any prompt (as contrasted with requiring task-specific training).

## Appendix Table A3. Publications describing prompt engineering automated optimization systems identified in a systematic search of literature

| Full citation |
| --- |
| Krishna, C. S.. Prompt Generate Train (PGT): A framework for few-shot domain adaptation, alignment, and uncertainty calibration of a retriever augmented generation (RAG) model for domain specific open book question-answering. Arxiv (2023). |
| Agarwal, Eshaan;Singh, Joykirat;Dani, Vivek;Magazine, Raghav;Ganu, Tanuja;Nambi, Akshay. PromptWizard: Task-Aware Prompt Optimization Framework. Arxiv (2024) . |
| Budagam, Devichand;Sankalp, K. J.;Kumar, Ashutosh;Jain, Vinija;Chadha, Aman. Hierarchical Prompting Taxonomy: A Universal Evaluation Framework for Large Language Models. Arxiv (2024) . |
| De Stefano, Gianluca;Schoenherr, Lea;Pellegrino, Giancarlo. Rag and Roll: An End-to-End Evaluation of Indirect Prompt Manipulations in LLM-based Application Frameworks. Arxiv (2024). |
| Hu, Silan;Wang, Xiaoning. FOKE: A Personalized and Explainable Education Framework Integrating Foundation Models, Knowledge Graphs, and Prompt Engineering. Arxiv (2024). |
| Liu, Dairui;Yang, Boming;Du, Honghui;Greene, Derek;Hurley, Neil;Lawlor, Aonghus;Dong, Ruihai;Li, Irene. RecPrompt: A Self-tuning Prompting Framework for News Recommendation Using Large Language Models. Arxiv (2024). |
| Nagle, Alliot;Girish, Adway;Bondaschi, Marco;Gastpar, Michael;Makkuva, Ashok Vardhan;Kim, Hyeji. Fundamental Limits of Prompt Compression: A Rate-Distortion Framework for Black-Box Language Models. Arxiv (2024). |
| Papadimitriou, Ioannis;Gialampoukidis, Ilias;Vrochidis, Stefanos. RAG Playground: A Framework for Systematic Evaluation of Retrieval Strategies and Prompt Engineering in RAG Systems. Arxiv (2024). |
| Pu, Xiao;He, Tianxing;Wan, Xiaojun. Style-Compress: An LLM-Based Prompt Compression Framework Considering Task-Specific Styles. Arxiv (2024). |
| Qin, Ruiyang;Ren, Pengyu;Yan, Zheyu;Liu, Liu;Liu, Dancheng;Nassereldine, Amir;Xiong, Jinjun;Ni, Kai;Hu, Sharon;Shi, Yiyu. NVCiM-PT: An NVCiM-assisted Prompt Tuning Framework for Edge LLMs. Arxiv (2024). |
| Habba, Eliya;Dahan, Noam;Lior, Gili;Stanovsky, Gabriel. PromptSuite: A Task-Agnostic Framework for Multi-Prompt Generation. Arxiv (2025). |
| Kriz, Anita;Janes, Elizabeth Laura;Shen, Xing;Arbel, Tal. Prompt4Trust: A Reinforcement Learning Prompt Augmentation Framework for Clinically-Aligned Confidence Calibration in Multimodal Large Language Models. Arxiv (2025). |
| Li, Haoyi;Yuan, Angela Yifei;Han, Soyeon Caren;Leckie, Christopher. SPADE: Systematic Prompt Framework for Automated Dialogue Expansion in Machine-Generated Text Detection. Arxiv (2025). |
| Li, Zongqian;Su, Yixuan;Collier, Nigel. PT-MoE: An Efficient Finetuning Framework for Integrating Mixture-of-Experts into Prompt Tuning. Arxiv (2025). |
| Liu, J.;Song, Y.;Jiang, R.;Li, Y.. Tasks-Embedded Reparameterization: A Novel Framework for Task-Specific Transfer Enhancement With Multitask Prompt Learning. International Journal of Intelligent Systems (2025) . |
| Liu, Jiang;Li, Bolin;Li, Haoyuan;Lin, Tianwei;Zhang, Wenqiao;Zhong, Tao;Yu, Zhelun;Wei, Jinghao;Cheng, Hao;He, Wanggui;Shu, Fangxun;Jiang, Hao;Lv, Zheqi;Li, Juncheng;Tang, Siliang;Zhuang, Yueting. Boosting Private Domain Understanding of Efficient MLLMs: A Tuning-free, Adaptive, Universal Prompt Optimization Framework. Arxiv (2025). |
| Nakada, Ryumei;Ji, Wenlong;Cai, Tianxi;Zou, James;Zhang, Linjun. A Theoretical Framework for Prompt Engineering: Approximating Smooth Functions with Transformer Prompts. Arxiv (2025). |
| Narayanaswamy, S. K.;Muniswamy, R.. Achieving Efficient Prompt Engineering in Large Language Models |